\documentclass[conference]{IEEEtran}

\usepackage[acronym]{glossaries}
\usepackage{cite}
\usepackage{amsmath,amssymb,amsfonts}
\usepackage{algorithmic}
\usepackage{graphicx}
\usepackage{textcomp}
\usepackage{xcolor}
\usepackage{booktabs}
\usepackage{hyperref}
\usepackage{url}

\def\BibTeX{{\rm B\kern-.05em{\sc i\kern-.025em b}\kern-.08em
    T\kern-.1667em\lower.7ex\hbox{E}\kern-.125emX}}

\newacronym{mpc}{MPC}{Model Predictive Control}
\newacronym{gnn}{GNN}{Graph Neural Network}
\newacronym{cnn}{CNN}{Convolutional Neural Network}
\newacronym{mse}{MSE}{Mean Squared Error}
\newacronym{mae}{MAE}{Mean Absolute Error}
\newacronym{lpips}{LPIPS}{Learned Perceptual Image Patch Similarity}
\newacronym{ssim}{SSIM}{Structural Similarity}
\newacronym{psnr}{PSNR}{Peak Signal-to-Noise Ratio}
\newacronym{vqa}{VQA}{Visual Question Answering}
\newacronym{mlp}{MLP}{Multi Layer Perceptron}
\newacronym{vae}{VAE}{Variational Autoencoder}
\newacronym{pcg}{PCG}{Procedural Content Generation}
\newacronym{savi}{SAVi}{Slot Attention for Video}
\newacronym{slam}{SLAM}{Simultaneous Localization And Mapping}
\newacronym{paig}{PAIG}{Physics-As-Inverse-Graphics}
\newacronym{lstm}{LSTM}{Long Short-Term Memory}
\newacronym{pde}{PDE}{Partial Differential Equation}
\newacronym{rkn}{RKN}{Recurrent Kalman Filter}
\newacronym{gru}{GRU}{Gated Recurrent Unit}
\newacronym{elu}{ELU}{Exponential Linear Unit}

\newif\iffinal
\finaltrue

\newif\ifappendix
\appendixfalse

\begin{document}

\title{ViPro-2: Unsupervised State Estimation via Integrated Dynamics for Guiding Video Prediction}

\iffinal
    \author{\IEEEauthorblockN{Patrick Takenaka}
    \IEEEauthorblockA{
            \textit{Stuttgart Media University}\\
            Stuttgart, Germany \\
            takenaka@hdm-stuttgart.de
            }
\and
\IEEEauthorblockN{Johannes Maucher}
\IEEEauthorblockA{
            \textit{Stuttgart Media University}\\
            Stuttgart, Germany \\
            maucher@hdm-stuttgart.de
            }
\and
\IEEEauthorblockN{Marco F. Huber}
\IEEEauthorblockA{
\textit{University of Stuttgart \emph{and} Fraunhofer IPA} \\
Stuttgart, Germany \\
            marco.huber@ieee.org
            }
            }
\fi

\maketitle

\begin{abstract}
Predicting future video frames is a challenging task with many downstream applications. Previous work~\cite{takenakaGuidingVideoPrediction2023, takenakaViProEnablingControlling2024} has shown that procedural knowledge enables deep models for complex dynamical settings, however their model ViPro assumed a given ground truth initial symbolic state. We show that this approach led to the model learning a shortcut that does not actually connect the observed environment with the predicted symbolic state, resulting in the inability to estimate states given an observation if previous states are noisy. In this work, we add several improvements to ViPro that enables the model to correctly infer states from observations without providing a full ground truth state in the beginning. We show that this is possible in an unsupervised manner, and extend the original Orbits dataset with a 3D variant to close the gap to real world scenarios.
\end{abstract}


\section{Introduction}
A large focus of current deep learning research is on the exploitation of ever increasing amounts of data. This approach has led to significant progress in recent years, culminating in breakthrough models such as ChatGPT~\cite{radfordImprovingLanguageUnderstanding} that made AI even more present in everyday life than ever before. However, as more and more parts of society rely on such models, questions are posed how robust and trustworthy such models are. Purely data-driven models give the impression of having a good understanding of the domain they were trained on, but many have shown that this is only superficial with limited capabilities for reasoning or extrapolation~\cite{marcusRebootingAIBuilding2019,mirzadehGSMSymbolicUnderstandingLimitations2024,duanGTBenchUncoveringStrategic2024}.
At the same time, domains where data is scarce do not benefit from this direction as much.

Informed machine learning~\cite{vonruedenInformedMachineLearning2023} aims to reduce these issues by leveraging domain knowledge in various forms to augment data-driven models. In this respect, Takenaka et al.~\cite{takenakaViProEnablingControlling2024} showed with their model ViPro the potential that procedural knowledge has to enhance existing deep learning models for video prediction, i.e., knowledge that represents the ``how-to'' of a given task (e.g., objects move according to a given equation) as opposed to declarative knowledge such as domain facts, describing the ``what''. This has the benefit of being completely generalizable for the domain of the integrated knowledge. For instance, a mathematical equation will be correct for any inputs that lie in its domain, and this behaviour is achieved implicitly. This is in contrast to gathering samples for a large amount of possible inputs that a purely data-driven model would need for this level of robustness.

ViPro focused on the integration of procedural knowledge in the form of dynamics equations for video prediction. They showed that scenes involving complex dynamics can be handled with this augmentation, while purely data-driven models struggle. 

However, they assumed a grounding of the model given the initial true dynamics state, and as we will show, this led to the model learning a shortcut where the true state is simply passed from frame to frame instead of being extracted from the observed video frame. While ViPro was still successful in predicting and decoding future frames, as soon as there are inaccuracies in the provided initial state, the model failed to correct and therefore predict the state accordingly. In this work, we restructure the learning task in a way that makes the model actually learn the state retrieval from the image domain, allowing more robustness in the face of incomplete starting states. We then show that we can train this model also with reduced initial conditioning, since the state estimation is robust enough. We further extend the Orbits dataset used in ViPro~\cite{takenakaViProEnablingControlling2024}, which involved only 2D dynamics, to 3D, to close the gap to real world applications.

The rest of this work is structured as follows: First, we describe related work in Section \ref{sec:related} and continue by describing ViPro and the aforementioned shortcomings in detail in Section \ref{sec:background}. We then offer our solution in Section \ref{sec:method} and show through a detailed ablation study how each proposed modification affects the result. Afterwards, we discuss how depth information can be utilized to handle 3D dynamics as well in Section \ref{sec:orbits3d}. Code and data is available at \url{https://github.com/P-Takenaka/ijcnn2025-vipro2}.

\section{Related Work}\label{sec:related}

\textbf{Dynamics Informed Video Prediction.} Video Prediction is a task that inherently demands a thorough understanding of the environment from the model. Here, the focus is not (only) on the visual plausibility and content diversity as is important for video generation tasks, but on the accurate modelling of the underlying domain processes. Doing this purely data-driven is challenging due to the often complex spatio-temporal relationships in the scene, and as such utilizing domain knowledge is a potential solution. Von Rueden et al.~\cite{vonruedenInformedMachineLearning2023} show a good overview of informed machine learning in general and introduce a taxonomy based on the type of knowledge being integrated. The integration of differential equations or other inductive biases based on physics knowledge in particular are of special interest for video prediction tasks, since the main goal is commonly the focus on the physical plausibility of the observed real-world scene~\cite{finnUnsupervisedLearningPhysical2016,fragkiadakiLearningVisualPredictive2016,fraccaroDisentangledRecognitionNonlinear2017,jannerReasoningPhysicalInteractions2018,kandukuriPhysicalRepresentationLearning2022}. Some approaches only utilize approximations of the underlying differential equations by relying on numerical integration methods such as the Euler method to obtain future state predictions~\cite{jaquesPhysicsasInverseGraphicsUnsupervisedPhysical2019, takenakaGuidingVideoPrediction2023,takenakaViProEnablingControlling2024}. These allow a broader applicability of the method if the exact solution of the underlying system of differential equations is unknown. Our work continues this line of research by improving the performance and applicability of previous methods~\cite{takenakaViProEnablingControlling2024}.

\textbf{Implicit State Estimation in Videos.} The most straightforward approach for estimating the visual scene state is through na\"ive state supervision. However, for real-world scenes accurate state labeling is very challenging. Alternatively, one can introduce sufficient inductive biases in the architecture and learning process to make the correct state prediction an emergent behaviour of the learning task. A common approach is the use of differentiable renderers~\cite{murthyGradSimDifferentiableSimulation2020,liangENVIDRImplicitDifferentiable2023}, which transform a symbolic, structured state into the image domain in a differentiable manner. This inductive bias in the state-to-image decoding then facilitates convergence to the true state that is consistent with the observed scene. However, their limitation is that the whole scene needs to be described in the symbolic state, thus limiting the applicability and generalization capability. Takenaka et al.~\cite{takenakaViProEnablingControlling2024} proposed an indirect state estimation scheme that was made possible by adding inductive biases already before the decoding step, while keeping a standard image decoder without specific inductive biases. This allowed taking into account latent information in the decoding process as well. However, their approach required the initial ground truth scene state, and as we show and remedy in this work, the state estimation process relied too much on this conditioning and fails when it is not available.

\section{Problem Statement}\label{sec:background}
Autoregressive video prediction assumes that a set of initial video frames is given, and the goal is to continue this sequence into the future. By providing more than a single initial frame, the model is able to observe not only the scene statics, but also the dynamics, including for instance object velocities or lighting effects. For complex dynamics, this is very challenging even when provided very long sequences, and one solution is to give the model information about the occurring dynamics. ViPro~\cite{takenakaViProEnablingControlling2024} showed how such procedural knowledge in the form of dynamics equations can assist a deep learning model in combining frame reconstructions with correct future frame predictions. 

\begin{figure}[t]
\begin{center}
    \includegraphics[trim={1.5cm 0 1.5cm 0},width=\linewidth,clip]{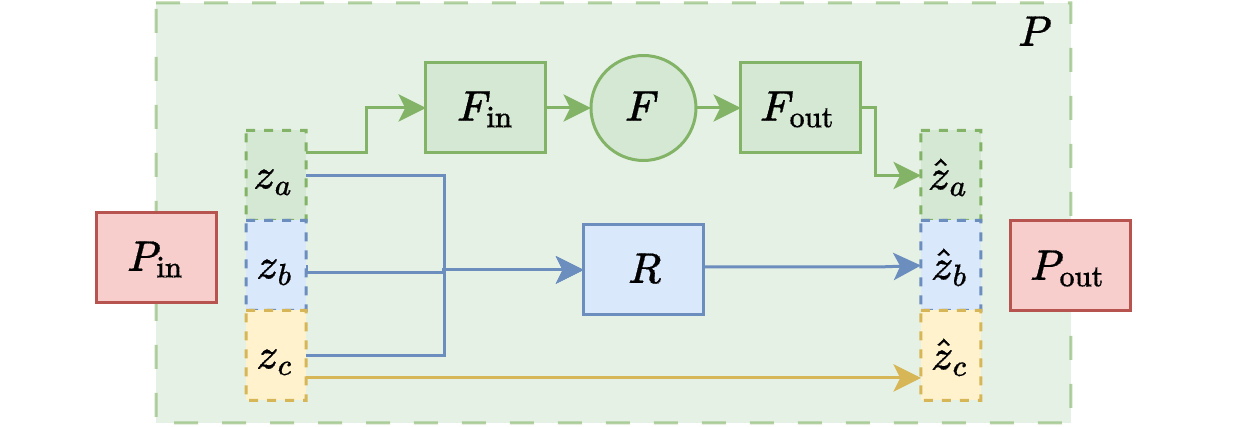}
\caption{Structure of the \emph{procedural knowledge module} $P$ of ViPro. The full latent vector $z$ is obtained via three distinct pathways that each handle different aspects of the scene to produce a full scene representation. $z_a$ (green pathway) corresponds to scene dynamics and as such information that is obtained from the integrated function $F$, $z_b$ (blue pathway) contains residual scene dynamics that are not handled by $F$ and is calculated using a transformer-based model $R$, and $z_c$ (yellow pathway) represents a shortcut connection modelling scene statics. Image courtesy of Takenaka et al.~\cite{takenakaViProEnablingControlling2024}.}
    \label{fig:procedural_knowledge_module}
\end{center}
\end{figure}

Their approach is to consider the integrated knowledge as a separate module in the overall architecture, and optimizing for video prediction then led to the model learning how to utilize this function to accomplish the task. For their object-centric architecture, they leveraged Slot Attention~\cite{locatelloObjectCentricLearningSlot2020} and the training scheme of \gls*{savi}~\cite{kipfConditionalObjectCentricLearning2022,elsayedSAViEndtoEndObjectCentric2022} and SlotFormer~\cite{wuSlotFormerUnsupervisedVisual2023} to separate the given scene into several latent object representations, from which the symbolic object state necessary for the integrated dynamics equations was extracted. During the \emph{burn-in phase}, i.e., the processing of the given initial video frames, the model operated a \emph{corrector model} (Slot Attention) and a separate \emph{predictor model} including the integrated dynamics equations---referred to as the \emph{procedural knowledge module} $P$ (cf. Figure \ref{fig:procedural_knowledge_module})---in tandem in order to sequentially process each frame. First, the corrector uses the observed (and encoded) video frame to adjust the current latent state, after which the predictor model produces a latent state for the next video frame. In the next frame iteration, this predicted latent state is then again modified by the corrector model given the next observation, and so on. The choice of Slot Attention for the corrector model also allows the application to object-centric scenes, since the latent frame representation is separated into per-object representations called slots.

\begin{figure*}[t]
\begin{center}
    \includegraphics[width=\linewidth]{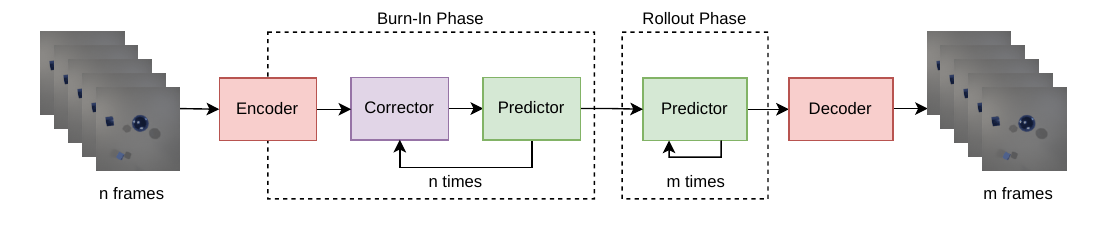}
\caption{General structure of the object-centric video prediction scheme of ViPro and the proposed ViPro-2 model. The model builds up an internal latent scene representation given $n$ burn-in frames by iteratively refining it via corrector and predictor models. Afterwards, the same predictor model is used to rollout future $m$ frames autoregressively. An encoder and a decoder model convert between this latent representation and the image domain.}
    \label{fig:video_prediction}
\end{center}
\end{figure*}

During the subsequent \emph{rollout phase}, i.e., when the model has to predict a video frame without a given observation, only the predictor model was used to produce a new latent video frame in each iteration. Figure \ref{fig:video_prediction} shows an overview of this video prediction process. For the very first frame during the burn-in phase, an initial latent state has to be provided, since there is no previous prediction step output that the corrector can adjust. In ViPro this initial state is constructed by encoding the ground truth symbolic state of each object. 

Takenaka et al.~\cite{takenakaViProEnablingControlling2024} showed that the model then successfully connects the available symbolic state of each object with the overall visual aspects in the scene to produce accurate frame predictions, and control over frame predictions was possible by simply modifying the symbolic state before decoding. However, while the transformation from symbolic state to video frame was possible, the reverse of obtaining a symbolic state from the video frame was not, as we will show. Since the model was provided the initial ground truth state and the integrated dynamics equations in $F$ described the scene dynamics in full, only a short-cut was learned that focused on carrying the symbolic ground truth state through the dynamics equations without considering the observed video frame. 

To highlight this, we test ViPro in a setting where the state initialization does not entirely happen from the ground
truth. Takenaka et al.~\cite{takenakaViProEnablingControlling2024} introduced the object-centric Orbits dataset, which features free moving objects that exert a gravitational pull on each other, resulting in chaotic, non-linear movements. As shown in their work, current video predictors struggle to predict the scene dynamics correctly in this setting, which makes it an ideal candidate for testing dynamics-augmented models. Instead of providing the model with the initial object world positions for this setting\footnote{We use a variant of the Orbits setting where each object is at the same distance from the camera. Otherwise, without providing an initial ground truth, depth estimation would be ambiguous without additional information due to varying object sizes. In Section \ref{sec:orbits3d} we approach this issue.}, we only provide screen coordinates instead. We then add an \gls*{mlp} model that gives the model the capacity to transform these screen coordinates back into a 3D world position---albeit without any supervision. We evaluate the image reconstruction performance based on the \gls*{lpips}~\cite{zhangUnreasonableEffectivenessDeep2018}---a metric utilizing learned feature embeddings that has shown to better capture human perception of similarity in an image compared to commonly used metrics such as \gls*{ssim} or \gls*{psnr}---and the state prediction performance by measuring the \gls*{mae} from the ground truth for the object positions. As can be seen in Table \ref{tab:screen_init_fail}, the model now struggles significantly and is unable to make use of the integrated knowledge. 

\begin{table}[t]
    \centering
    \caption{Comparison of ViPro performance of initialization of the symbolic state with the full ground truth state and a reduced conditioning variant that only uses the screen coordinates instead of the 3D position.}
    \begin{tabular}{l|l|l}
         \textbf{State Initialization} & \textbf{LPIPS}$\downarrow$ & \textbf{Position MAE}$\downarrow$ \\\midrule
         Full Ground Truth & $3.4$\textcolor{darkgray}{\scriptsize$\pm0.1$} & $0.18$\textcolor{darkgray}{\scriptsize$\pm0.0$}\\
         Screen Coordinates & $31.8$\textcolor{darkgray}{\scriptsize$\pm0.3$} & $9.54$\textcolor{darkgray}{\scriptsize$\pm0.7$}\\
    \end{tabular}
    \label{tab:screen_init_fail}
\end{table}

This is surprising, since in a setting with only 2D dynamics the screen coordinate should still be sufficiently informative to estimate true world positions from it. When providing ground truth as the initial state, the model has in fact two learning directions for estimating the future state: Either---as the authors of ViPro intended---it finds a correlation between the current symbolic state and the currently observed video frame, and utilizes this relationship to estimate the symbolic state from the video, or---as we argue is more likely---the model simply passes the initial ground truth state from frame to frame without considering the observed video. This is possible because the integrated function is perfectly describing the underlying dynamics, and---if the model learns this simple relationship---results in a sequence of ground truth symbolic states for each video frame that the model then only needs to decode back into an image.

We argue that ViPro follows the latter learning directions as indicated by the inability to correctly utilize the screen coordinates, since this requires further processing before it is usable by the integrated function.

\section{Proposed Solution}\label{sec:method}
Our new architecture---termed ViPro-2---builds on top of ViPro and introduces several modifications and improvements to the original method that allows true joint state estimation and video prediction and even surpasses the performance of ViPro with the initial ground truth state. Table \ref{tab:model_transition} summarizes the performance gains that each change brings with it and shows the gradual performance transition from the ViPro model to our final ViPro-2 model. In the following we explain each modification in more detail. We group these modifications into four stages that also serve as an ablation study to identify, which changes contributed how much to the overall performance gain. We grouped changes into a single stage when we observed that only their joint application lead to an overall performance increase.

\begin{table}[t]
    \centering
    \caption{From top to bottom, the transition from ViPro to our proposed ViPro-2 model, where each row includes the changes from above rows. The symbolic state is initialized from the screen coordinates instead of the 3D world coordinates.}
    \begin{tabular}{p{4.2cm}|p{1.3cm}|p{1.3cm}}
        \textbf{Configuration} & \textbf{LPIPS}$\downarrow$ & \textbf{MAE}$_{Pos}\downarrow$ \\\midrule
         ViPro & $31.8$\textcolor{darkgray}{\scriptsize$\pm0.0$} & $9.54$\textcolor{darkgray}{\scriptsize$\pm0.7$} \\\midrule
         (A) Restructuring $P$ & $26.5$\textcolor{darkgray}{\scriptsize$\pm2$} & $9.94$\textcolor{darkgray}{\scriptsize$\pm0.1$}\\
         (B) Observation Alignment & $3.8$\textcolor{darkgray}{\scriptsize$\pm0.2$} & $0.49$\textcolor{darkgray}{\scriptsize$\pm0.04$} \\
         (C) Separated Latent State & $3.2$\textcolor{darkgray}{\scriptsize$\pm0.3$} & $0.35$\textcolor{darkgray}{\scriptsize$\pm0.0$} \\
         \midrule
         \textbf{(D) Gain Predictor (ViPro-2)} & $\mathbf{1.72}$\textcolor{darkgray}{\scriptsize$\pm0.05$} & $\mathbf{0.042}$\textcolor{darkgray}{\scriptsize$\pm0.0$} \\ 
    \end{tabular}
    \label{tab:model_transition}
\end{table}

We train the models with mostly the same approach as in ViPro. The main difference is that we only train with six unroll frames (twelve for ViPro) as we found that this does not hinder validation performance, while reducing the training times. The number of validation burn-in and unroll frames remain the same as in ViPro at six and 24, respectively, making sure that long-term prediction capabilities are measured in the same manner.

\subsection{Restructuring P}

In ViPro, the \emph{procedural knowledge module} $P$ is responsible for merging the symbolic state with the latent state in order to produce suitable latent representations for decoding afterwards. The modules $P_{\mathrm{in}}$ and $P_{\mathrm{out}}$ included non-linearities and transformed the latent representation of the underlying \gls*{savi} model into a separable latent representation, after which the linear modules $F_{\mathrm{in}}$ and $F_{\mathrm{out}}$ handled the transformation of this latent space to the symbolic space. We observed that this separation is not necessary, and is even detrimental to image reconstruction performance if the estimated symbolic state has diverged from the ground truth. Instead, we omit both $P_{\mathrm{in}}$ and $P_{\mathrm{out}}$ and instead include non-linearities in $F_{\mathrm{in}}$ and $F_{\mathrm{out}}$ by realizing them as single layer \glspl*{mlp} with ReLU activations. In Table \ref{tab:model_transition} we can see that this slightly improves image prediction performance, but not to an acceptable level yet. Still, this serves as a foundation for the following changes.

\subsection{Observation Alignment}\label{sec:stage_2}

The training process of ViPro optimizes for both the image reconstruction through the reconstruction loss $\mathcal{L}_\mathrm{rec}$ and an auxiliary auto-encoder loss $\mathcal{L}_\mathrm{AE}$, the latter informing the model that $F_{\mathrm{in}}$ and $F_{\mathrm{out}}$ should represent a bijective relationship according to
\begin{equation}
    \mathcal{L} = \mathcal{L}_\mathrm{rec} + \mathcal{L}_\mathrm{AE}~.
\end{equation}

However, convergence of the symbolic state to the ground truth is only achieved through the indirect relationship with the image reconstruction performance, where good symbolic state predictions should correlate with the movements in the image, and as such lead to better image predictions. This, however, does not seem to be sufficient, which is indicated by a large \gls*{mae} for the predictions of the object positions. While state supervision would be an obvious solution, we found that we can also achieve better symbolic state predictions by adding another auxiliary loss that does not need ground truth, which we call the observation loss $\mathcal{L}_\mathrm{obs}$.

Inspired by recursive state estimators such as Kalman filters, we want to consider both the current symbolic state observation $s_\mathrm{obs}$---coming from our image encoder---and the symbolic state prediction $s_\mathrm{pred}$ of $F$ from the last time step when making the next state prediction. 

\newpage

We then define $\mathcal{L}_\mathrm{obs}$ as

\begin{equation}
    \mathcal{L}_\mathrm{obs} = (s_\mathrm{pred} - s_\mathrm{obs})^2
\end{equation}
so that its minimization results in an alignment of $s_\mathrm{pred}$ and $s_\mathrm{obs}$.
The combined loss therefore is

\begin{equation}
    \mathcal{L} = \mathcal{L}_\mathrm{rec} + \mathcal{L}_\mathrm{AE} + \mathcal{L}_\mathrm{obs}~.
\end{equation}

We did not observe any benefit in weighing the loss terms differently and as such omit weighting factors in the equation.

\begin{figure*}[t]
\begin{center}
    \includegraphics[width=.9\textwidth]{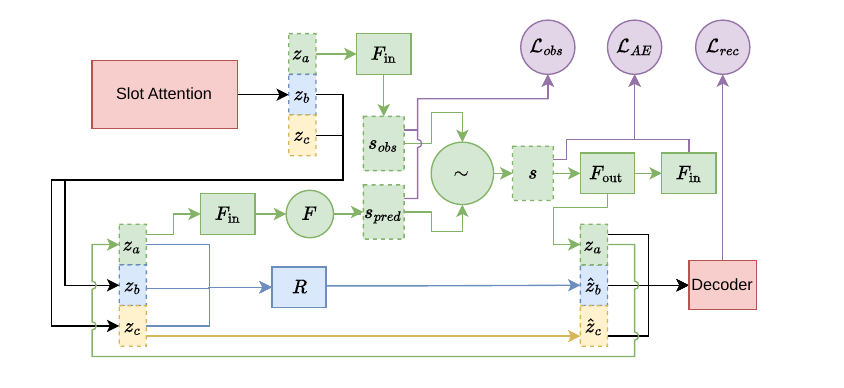}
\caption{Structure of our new procedural knowledge module in ViPro-2. The latent vector $z_a$ (\textcolor{green}{green} pathways) is produced by considering both the observed state $s_{\mathrm{obs}}$ from the current burn-in frame and the predicted state $s_\mathrm{pred}$ coming from the integrated dynamics equations in $F$. As we show in our ablations this fusion ($\sim$ node in the figure) can in principle be realized by various techniques, however we found the best way to be a recurrent gain predictor model $G$ (cf. Section \ref{sec:stage_4}). The diagram also shows the input sources for the various losses used in training (\textcolor{purple}{purple}), namely the image reconstruction loss $\mathcal{L}_\mathrm{rec}$, the auto-encoding loss $\mathcal{L}_\mathrm{AE}$, and our new observation alignment loss $\mathcal{L}_{\mathrm{obs}}$ (cf. Section \ref{sec:stage_2}).}
    \label{fig:new_procedural_knowledge_module}
\end{center}
\end{figure*}

To further force the model to actually utilize the observation when making the symbolic state prediction we only use $s_\mathrm{obs}$ when transforming the symbolic state to the latent state before decoding. $s_\mathrm{pred}$ is as such only used as a proxy target state. Figure \ref{fig:new_procedural_knowledge_module} shows the overall structure of our new \emph{procedural knowledge module} $P$ including the various losses used in training. For comparison, we also show the module $P$ as used in ViPro in Figure \ref{fig:procedural_knowledge_module}.

This change in the training target resulted in a significant performance improvement for both the image reconstruction, and also the state prediction. Still, in contrast to the initialization from the ground truth of ViPro in Table \ref{tab:screen_init_fail}, the performance lags behind. This indicates that the model continues to have difficulty in correctly connecting the symbolic state with the observed frame.

\subsection{Separated Latent State}\label{sec:stage_3}

To improve the performance further, we now introduce a set of sensible modifications to the architecture with the aim of making the convergence easier for the model without losing on generalization capability. 

First, we subdivide the latent state $z_a$ into separate vectors for each individual variable (e.g., position and velocity in the Orbits setting) and use separate \glspl*{mlp} for each within $F_\mathrm{in}$ and $F_\mathrm{out}$. The reasoning here is that these components only transform between symbolic and latent representation and not do state predictions, and as such each variable's transformation should not be influenced by the transformation of other variables. 

We then make use of this disentangled state and only supply \textit{observable} variables to the decoder model that reconstructs the images. We consider a variable to be \textit{observable} if a single video frame is sufficient to make a meaningful prediction---in the Orbits setting this would be the object positions, but not the object velocities. We hypothesize that since our decoder decodes video frames individually, it should not have any use of \textit{unobservable} variables such as the object velocities. While in theory the model could learn this relationship on its own, we have found that this change improves convergence.

Finally, we also no longer apply the auto-encoder loss $\mathcal{L}_\mathrm{AE}$ and observation loss $\mathcal{L}_\mathrm{obs}$ to the background slot state. The object-centric slots produced by the underlying Slot Attention have to represent the whole scene, and since not all areas in the frames correspond to an object, one slot gravitates towards representing the background. So far in ViPro, the background slot is not handled any different than the foreground slots---a fact that most likely results in a more complex learning problem for the model, which we remedy in the next set of changes in Section \ref{sec:stage_4}. As such, since the background does not represent an object, the losses that focus on correct object state convergence should not apply to it. Interestingly we found that this change was now necessary with the subdivided $z_a$ to produce good results, whereas previously it would not be beneficial. We hypothesize that due to the now cut connections between variables in $z_a$, the model is lacking capacity to correctly establish the background state, and the additional regularization through $\mathcal{L}_\mathrm{obs}$ and $\mathcal{L}_\mathrm{AE}$ is constraining the model too much for the background.

We note that using the changes in this stage individually resulted in a performance decrease, and only their combination lead to a better overall performance, which is why we grouped them here.

\subsection{Gain Predictor}\label{sec:stage_4}

Our final set of changes first addresses the issues of background and foreground slot states that was mentioned in Section \ref{sec:stage_3}, and then improves on the state fusion of $s_\mathrm{obs}$ and $s_\mathrm{pred}$. As mentioned before, in ViPro the background slot is handled in the same manner as the foreground slots. This means that in each iteration $F_\mathrm{out}$ produces a latent state $z_a$ for the background slot as well, in essence describing its dynamics state of position and velocity in the Orbits setting---a notion that does not make much sense for a non-object. This is also true for the slot initialization, where the initial conditioning for the background slot is simply a zero vector, and the model has to correctly learn to interpret this as not being an  object. While this has worked sufficiently well for ViPro, we found that this seems like an unnecessary hack that increases the learning complexity for the model. As such, we instead completely separate the background slot state from the foreground slots and no longer modify it through $F_\mathrm{out}$. Furthermore, the initial state of the background slot is now a learned vector initialized from a normal distribution, which gives the model the flexibility to adjust the background state as it deems necessary during training.

In Section \ref{sec:stage_2}, we found that a separation of state estimation into prediction $s_\mathrm{pred}$ and observation $s_\mathrm{obs}$ has led to a significant performance increase. However, as described we use $s_\mathrm{pred}$ only as a proxy target for $\mathcal{L}_\mathrm{obs}$. In methods such as Kalman filters this is usually done more intelligently, where a gain factor $K \in [0,1]$ is used to weigh between prediction and observation to obtain the final state $s$, as in

\begin{equation}
    s = s_\mathrm{pred} + K (s_\mathrm{obs} - s_\mathrm{pred})~.
\end{equation}

In Section \ref{sec:stage_2}, we effectively set $K=1$ leading to $s = s_\mathrm{obs}$, with $s_\mathrm{pred}$ only acting as regularizer and not being available for the estimation. If $K$ is instead dynamically calculated for each frame based on the (past) observations and predictions, the model has the option to take into account the observation and prediction noise when determining $K$.

We implement a recurrent gain predictor $G$ which takes concatenated $z_{a_\mathrm{pred}}$ and $z_{a_\mathrm{obs}}$ as inputs and outputs $K$. It consists of an initial non-linear \gls*{mlp}, followed by a \gls*{gru} \cite{choLearningPhraseRepresentations2014}, and a final non-linear \gls*{mlp}. We join the individual object features into a single dimension in $G$ to be able to model inter-object relations. 

We finally transform the output of $G$ with the sigmoid function $\sigma$ to obtain $K \in [0,1]$ according to

\begin{equation}
    K = \sigma(G(z_{a_\mathrm{obs}}, z_{a_\mathrm{pred}}))~.
\end{equation}

We show the performance of this model in Table \ref{tab:gain_prediction_results}, together with the best performing model of Section \ref{sec:stage_3}, which corresponds to a model with a constant $K=1.0$ in the new state estimation scheme. For reference we also compare with a variant model with a constant $K=0.5$ to better highlight the utility of a dynamic $K$.

\begin{table}[t]
    \centering
    \caption{Comparison of model performance for different strategies of estimating the gain factor $K$.}
    \begin{tabular}{l|l|l}
         \textbf{State Initialization} & \textbf{LPIPS}$\downarrow$ & \textbf{Position MAE}$\downarrow$ \\\midrule
         Default ($K=1.0$) & $5.0$\textcolor{darkgray}{\scriptsize$\pm0.9$} & $0.498$\textcolor{darkgray}{\scriptsize$\pm0.1$} \\
         Average ($K=0.5$) & $6.9$\textcolor{darkgray}{\scriptsize$\pm7.5$} & $3.1$\textcolor{darkgray}{\scriptsize$\pm5.1$} \\
         Gain Predictor & $\mathbf{1.72}$\textcolor{darkgray}{\scriptsize$\pm0.1$} & $\mathbf{0.042}$\textcolor{darkgray}{\scriptsize$\pm0.0$} \\
    \end{tabular}
    \label{tab:gain_prediction_results}
\end{table}

\begin{figure}[t]
    \centering
    \includegraphics[width=\linewidth]{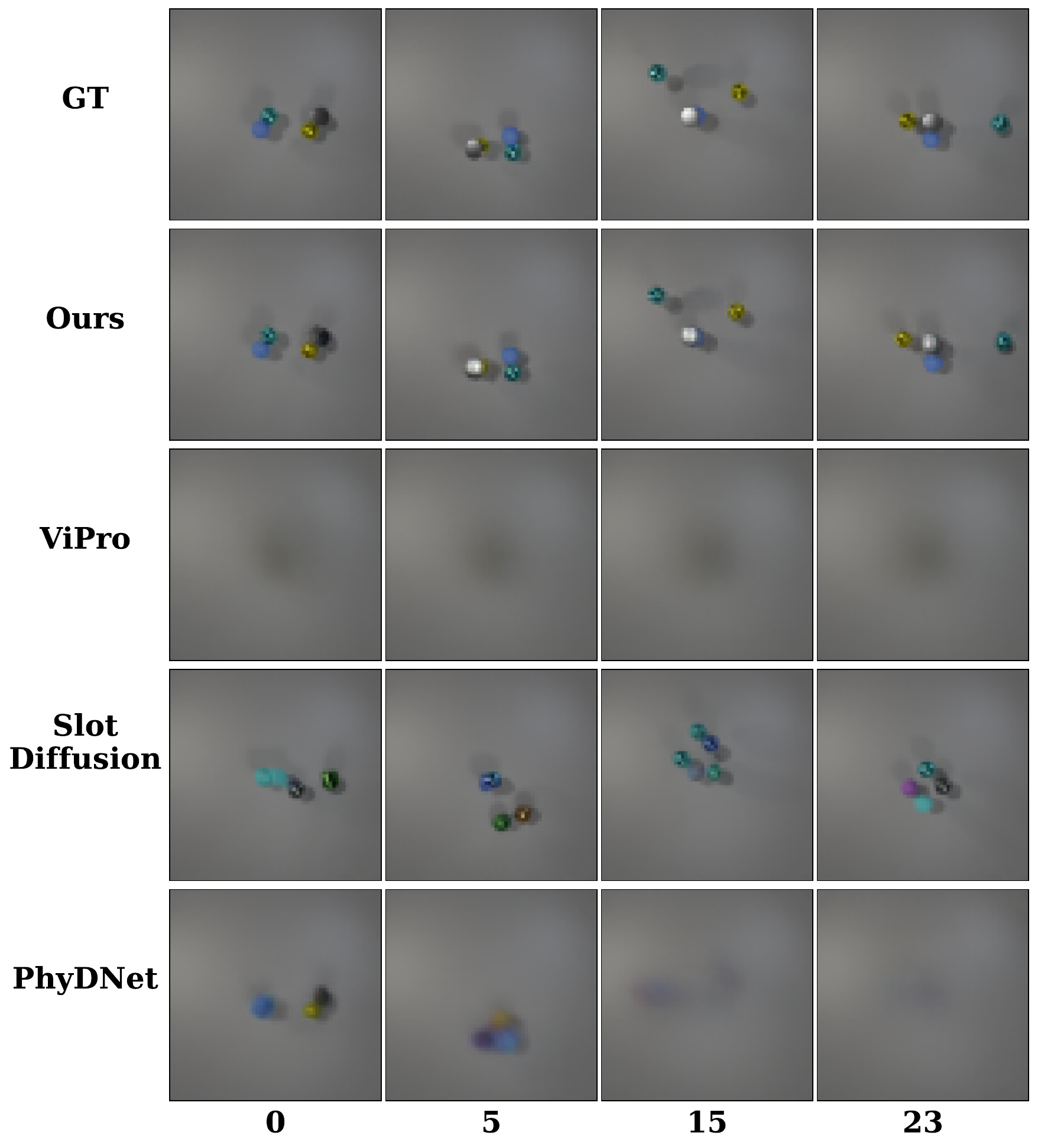}
    \caption{Qualitative performance of our model for the Orbits-2D setting compared to the ground-truth (GT) and other related work. Predictions of selected frame iterations are shown from left to right.}
    \label{fig:base_performance}
\end{figure}

We compare our final model with related work in Table~\ref{tab:base_performance} and visualize predictions in Figure~\ref{fig:base_performance}. Our observations align with Takenaka et al.~\cite{takenakaViProEnablingControlling2024}, in that other purely data-driven, but also physics-informed related work is unable to correctly predict frames for the Orbits dataset. In our case, however, the ViPro model also struggles with the reduction of initial state assumptions. A variant of our model in which we supervise the object states during the burn-in phase results in the same overall performance, and shows that even unsupervised we can achieve the same results.

\begin{table}[t]
    \centering
    \caption{Image reconstruction performance comparison of our proposed complete architecture and relevant related work for the Orbits-2D dataset.}
    \begin{tabular}{p{2.8cm}|p{1.3cm}|p{1.3cm}|p{1.3cm}}
                      & \textbf{LPIPS}$\downarrow$ & \textbf{SSIM}$\uparrow$ & \textbf{PSNR}$\uparrow$ \\\midrule
                    \textbf{ViPro-2 (Ours)} & $\mathbf{1.72}$\textcolor{darkgray}{\scriptsize$\pm0.1$} & $\mathbf{99.0}$\textcolor{darkgray}{\scriptsize$\pm0.0$} & $\mathbf{40.9}$\textcolor{darkgray}{\scriptsize$\pm0.1$}\\
                    \midrule
                    ViPro-2 Supervised & $1.72$\textcolor{darkgray}{\scriptsize$\pm0.0$} & $98.9$\textcolor{darkgray}{\scriptsize$\pm0.0$} & $40.2$\textcolor{darkgray}{\scriptsize$\pm0.1$} \\
                    \midrule
                    Slot Diffusion &  $15.3$\textcolor{darkgray}{\scriptsize$\pm0.0$} &  $83.0$\textcolor{darkgray}{\scriptsize$\pm0.0$} &  $27.0$\textcolor{darkgray}{\scriptsize$\pm0.0$} \\
                    ViPro & $31.8$\textcolor{darkgray}{\scriptsize$\pm0.0$} &   $86.6$\textcolor{darkgray}{\scriptsize$\pm0.0$} &  $28.5$\textcolor{darkgray}{\scriptsize$\pm0.0$}  \\
                    SlotFormer~\cite{wuSlotFormerUnsupervisedVisual2023}    &  $23.3$\textcolor{darkgray}{\scriptsize$\pm0.0$} &  $86.5$\textcolor{darkgray}{\scriptsize$\pm0.0$} &  $28.6$\textcolor{darkgray}{\scriptsize$\pm0.0$} \\
PhyDNet~\cite{leguenDisentanglingPhysicalDynamics2020}        &  $18.3$\textcolor{darkgray}{\scriptsize$\pm0.9$} &  $88.8$\textcolor{darkgray}{\scriptsize$\pm0.1$} &  $29.8$\textcolor{darkgray}{\scriptsize$\pm0.0$} \\
PredRNN-V2~\cite{wangPredRNNRecurrentNeural2023}     &  $18.3$\textcolor{darkgray}{\scriptsize$\pm0.5$} &  $89.2$\textcolor{darkgray}{\scriptsize$\pm0.2$} &  $30.0$\textcolor{darkgray}{\scriptsize$\pm0.1$} \\
Dona et al.~\cite{donaPDEDrivenSpatiotemporalDisentanglement2021}    &  $30.7$\textcolor{darkgray}{\scriptsize$\pm0.3$} &  $86.0$\textcolor{darkgray}{\scriptsize$\pm0.6$} &  $28.4$\textcolor{darkgray}{\scriptsize$\pm0.4$} \\
\end{tabular}
    \label{tab:base_performance}
\end{table}

\section{Moving to 3D}\label{sec:orbits3d}
The Orbits dataset of Takenaka et al.~\cite{takenakaViProEnablingControlling2024} and also in our previous section omitted the handling of depth by limiting the object dynamics to 2D. In order to extend the applicability we now introduce a scheme to also be able to handle 3D dynamics as well. We introduce a dataset variant termed Orbits-3D that now has objects moving in all three dimensions. Predicting the distance from the camera to each object is not well defined however. Since each object is of varying scale and can appear at various depths, the distance to the camera is ambiguous. Therefore we introduce a scheme that utilizes a depth map provided with the first video frame in order to facilitate the estimation of a good initial position. Since RGB-D cameras are getting more and more prevalent, we found this to be a realistic assumption. We have found that the best way to integrate the depth is directly in the slot initialization step instead of providing it to, for instance, the image encoder as an additional channel or even to an auxiliary input head. 

We utilize the available initial screen coordinate of each object to sample from the depth map, and use this sampled value together with the screen coordinate as input to an \gls*{mlp} that predicts the initial world positions. While this depth might not be ideal in every case---since objects are not single points but have varying shapes and sizes, resulting in different depths depending on which pixel we sample of the object---we found this to be a good enough approximation for the model to produce good frame predictions.

\begin{figure}[t]
    \centering
    \includegraphics[width=\linewidth]{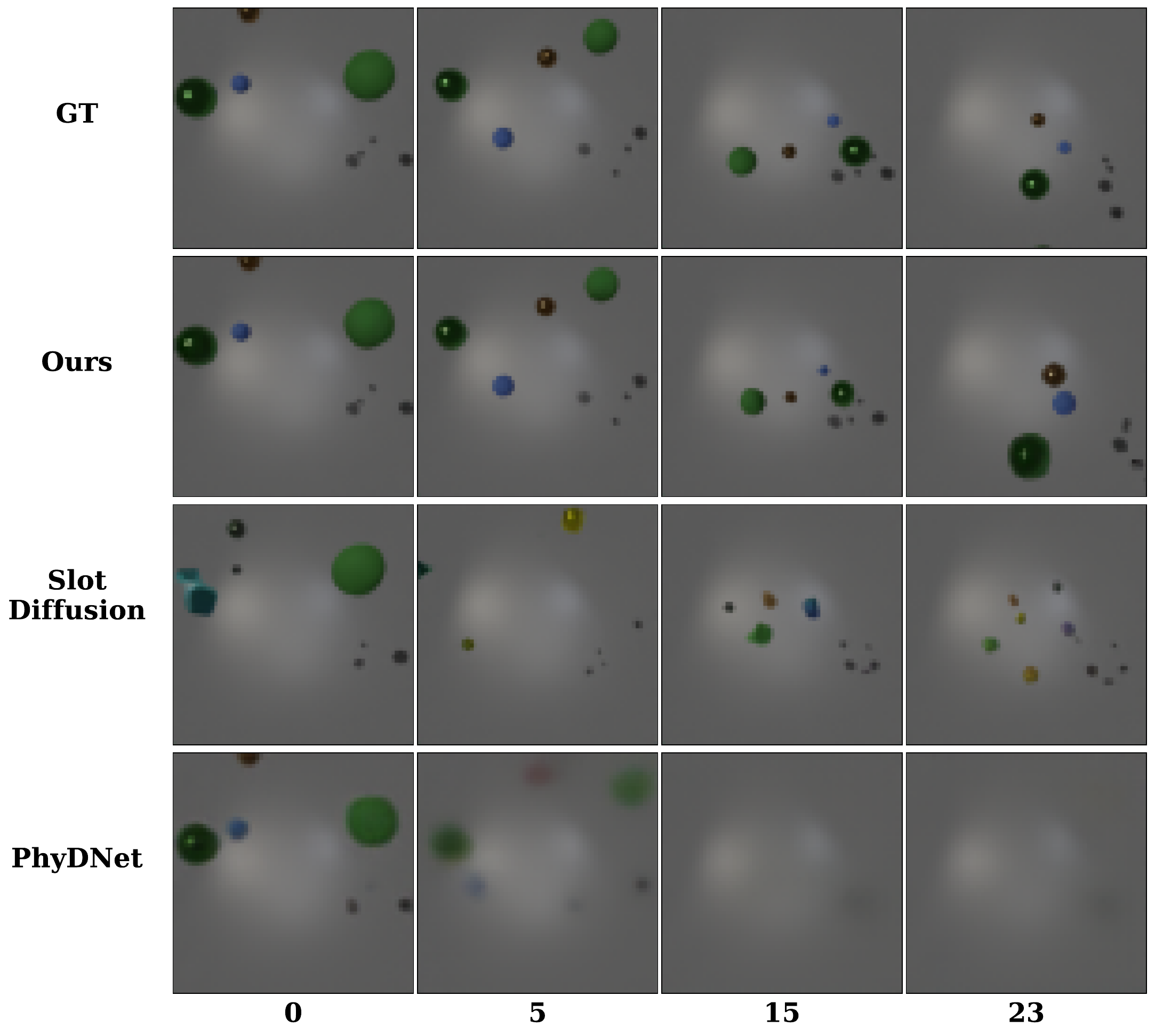}
    \caption{Qualitative performance of our model for the Orbits-3D setting compared to the ground truth (GT). Predictions of selected frame iterations are shown from left to right. Our model is still able to position objects correctly in future frames.}
    \label{fig:base_3d_performance}
\end{figure}

We also modified the reconstruction loss $\mathcal{L_\mathrm{rec}}$ for this dataset, where instead of using the \gls*{mse}, we now use the \gls*{mae} together with an \gls*{lpips}-based loss, same as is used in Slot Diffusion~\cite{wuSlotDiffusionObjectCentricGenerative2023}. We have found this to improve the performance in the 3D setting even further, however it was not helpful in the Orbits-2D setting, most likely because the frame predictions were already of sufficient quality for the loss to have a beneficial impact. Otherwise the model is kept the same as in Section \ref{sec:stage_4}. We show the results and ablations in Table \ref{tab:orbits_3d_loss_ablations} and Table \ref{tab:orbits_3d_depth_ablations} for depth integration methods and different loss terms respectively, and comparisons with other models in Table \ref{tab:orbits3d_performance} and Figure \ref{fig:base_3d_performance}.

\begin{table}[t]
    \centering
    \caption{Model performance for the Orbits-3D setting for different loss terms. For all configurations we use the ``Single Depth Sample'' setup from Table \ref{tab:orbits_3d_depth_ablations}.}
    \begin{tabular}{l|l|l}
         \textbf{State Initialization} & \textbf{LPIPS}$\downarrow$ & \textbf{Position MAE}$\downarrow$ \\\midrule
         MSE $\mathcal{L}_\mathrm{rec}$ (w/o depth) & $30.0$\textcolor{darkgray}{\scriptsize$\pm2.9$} & $31.56$\textcolor{darkgray}{\scriptsize$\pm0.9$} \\
         MSE $\mathcal{L}_\mathrm{rec}$ & $10.0$\textcolor{darkgray}{\scriptsize$\pm0.4$} & $\mathbf{14.9}$\textcolor{darkgray}{\scriptsize$\pm1.5$} \\
         LPIPS $\mathcal{L}_\mathrm{rec}$ & $7.1$\textcolor{darkgray}{\scriptsize$\pm0.2$} & $\mathbf{16.2}$\textcolor{darkgray}{\scriptsize$\pm2.9$}  \\
         MAE + LPIPS $\mathcal{L}_\mathrm{rec}$ & $\mathbf{6.9}$\textcolor{darkgray}{\scriptsize$\pm0.1$} & $\mathbf{15.7}$\textcolor{darkgray}{\scriptsize$\pm2.1$} \\
    \end{tabular}
    \label{tab:orbits_3d_loss_ablations}
\end{table}

\begin{table}[t]
    \centering
    \caption{Model performance for the Orbits-3D setting using different means of depth estimation. In ``Single Depth Sample'' we sample a depth value for each object using the screen coordinate as described in Section \ref{sec:orbits3d}. For ``Depth Map as Input'' we instead supply the depth map as a whole additional channel to the image encoder. For all configurations we use the loss ``MAE + LPIPS $\mathcal{L}_\mathrm{rec}$'' setup from Table \ref{tab:orbits_3d_loss_ablations}.}
    \begin{tabular}{l|l|l}
         & \textbf{LPIPS}$\downarrow$ & \textbf{Position MAE}$\downarrow$ \\\midrule
         Baseline (w/o depth) & $16.3$\textcolor{darkgray}{\scriptsize$\pm1.8$} & $31.9$\textcolor{darkgray}{\scriptsize$\pm2.4$} \\
         \midrule
         Single Depth Sample & $\mathbf{6.9}$\textcolor{darkgray}{\scriptsize$\pm0.1$} & $\mathbf{15.7}$\textcolor{darkgray}{\scriptsize$\pm2.1$} \\
         Depth Map as Input & $15.4$\textcolor{darkgray}{\scriptsize$\pm0.5$} & $33.6$\textcolor{darkgray}{\scriptsize$\pm0.8$} \\
         Depth Sample + Depth Map & $7.2$\textcolor{darkgray}{\scriptsize$\pm0.2$} & $19.1$\textcolor{darkgray}{\scriptsize$\pm5.2$} \\
         \midrule
         Groundtruth Z & $1.4$\textcolor{darkgray}{\scriptsize$\pm0.0$} & $0.08$\textcolor{darkgray}{\scriptsize$\pm0.0$} \\
    \end{tabular}
    \label{tab:orbits_3d_depth_ablations}
\end{table}

\begin{table}[t]
    \centering
    \caption{Image reconstruction performance comparison of our proposed complete architecture and relevant related work for the Orbits-3D dataset.}
    \begin{tabular}{p{2.8cm}|p{1.3cm}|p{1.3cm}|p{1.3cm}}
                      & \textbf{LPIPS}$\downarrow$ & \textbf{SSIM}$\uparrow$ & \textbf{PSNR}$\uparrow$ \\\midrule
                    \textbf{ViPro-2 (Ours)} & $\mathbf{6.9}$\textcolor{darkgray}{\scriptsize$\pm0.1$} & $\mathbf{88.9}$\textcolor{darkgray}{\scriptsize$\pm0.8$} & $\mathbf{28.1}$\textcolor{darkgray}{\scriptsize$\pm0.5$}\\
                    \midrule
                    ViPro-2 Supervised & $2.5$\textcolor{darkgray}{\scriptsize$\pm0.1$} & $95.9$\textcolor{darkgray}{\scriptsize$\pm0.3$} & $33.0$\textcolor{darkgray}{\scriptsize$\pm0.3$}\\
                    \midrule
                    Slot Diffusion &  $32.8$\textcolor{darkgray}{\scriptsize$\pm0.0$} &  $66.0$\textcolor{darkgray}{\scriptsize$\pm0.0$} &  $20.0$\textcolor{darkgray}{\scriptsize$\pm0.0$} \\
                    ViPro & $33.5$\textcolor{darkgray}{\scriptsize$\pm1.8$} &  $80.7$\textcolor{darkgray}{\scriptsize$\pm0.2$} &  $25.3$\textcolor{darkgray}{\scriptsize$\pm0.1$}  \\
                    SlotFormer  &  $46.1$\textcolor{darkgray}{\scriptsize$\pm0.8$} &  $70.7$\textcolor{darkgray}{\scriptsize$\pm0.8$} &  $23.4$\textcolor{darkgray}{\scriptsize$\pm0.4$} \\
PhyDNet     &  $26.3$\textcolor{darkgray}{\scriptsize$\pm0.8$} &  $82.2$\textcolor{darkgray}{\scriptsize$\pm0.3$} &  $25.5$\textcolor{darkgray}{\scriptsize$\pm0.0$} \\
PredRNN-V2  &  $24.7$\textcolor{darkgray}{\scriptsize$\pm0.3$} &  $83.4$\textcolor{darkgray}{\scriptsize$\pm0.3$} &  $26.3$\textcolor{darkgray}{\scriptsize$\pm0.1$} \\
Dona et al. &  $36.2$\textcolor{darkgray}{\scriptsize$\pm0.8$} &  $89.2$\textcolor{darkgray}{\scriptsize$\pm7.7$} &  $22.8$\textcolor{darkgray}{\scriptsize$\pm2.7$} \\
\end{tabular}
    \label{tab:orbits3d_performance}
\end{table}

The results indicate that only by modifying the reconstruction target we can get improvements for both image fidelity and also the state prediction. In contrast to the 2D setting, however, the position estimation error is still high. A large contributor to the overall error is the Z coordinate, which amounts to around 90\% of the total error. The comparatively low X/Y error combined with frame predictions of objects at the correct locations indicate, however, that the model is still able to work with this noisy state to produce useful outputs.

Still, it is likely that further improvements to the depth integration could lead to better performance. In order to test potential performance gains given a perfect depth estimation, we test our model by always providing the ground-truth $z$ coordinate of all objects, in addition to the screen coordinate (see experiment ``Groundtruth Z'' in Table \ref{tab:orbits_3d_depth_ablations}). Here, the position estimation error drops drastically, which also led to much better image reconstruction performance. As such, better performance can be achieved by improving the depth estimation further instead of resorting to doing supervision. 

\section{Limitations and Future Work}
In this work, we still did not assume a fully unconditioned model. In our experiments we have observed that slot attention has trouble discovering objects in our datasets when not conditioned properly---a better object discovery mechanism could help in this regard. Furthermore, we have not yet tackled the estimation of temporal states such as the object velocity, which would introduce another level of complexity, and doing so completely unsupervised will be very challenging. 

This line of research is also inherently limited by its foundational assumption that the underlying dynamics equations are available---although we argue that for some domains such as the Orbits setting no purely data-driven method proved feasible in our experiments so far. A middle-ground that can handle only partially available dynamics knowledge could be a hybrid approach that combines the integration with the synthesis of novel knowledge. 

A first step towards this could be system identification that aims to learn the parameters of the integrated function.

In the future it will be worth exploring the application of our approach to areas relying on deep visual scene understanding such as robotics or autonomous systems. Takenaka et al.~\cite{takenakaViProEnablingControlling2024} with their Pendulum Camera dataset have already shown the potential of this line of work for visual navigation, and further research could enable more accurate scene predictions based on the action policy and model of autonomous agents.

\section{Conclusion}
In this work we showed how the procedural knowledge guided video prediction model ViPro~\cite{takenakaViProEnablingControlling2024} can be made more robust towards incomplete conditioning data, and adjust its learning process in a way that enables it to truly estimate the state from the observations. We further extended the Orbits dataset introduced by Takenaka et al.~\cite{takenakaViProEnablingControlling2024} to 3D and demonstrate how depth information can be used to still be able to benefit from the integrated procedural knowledge.

\bibliographystyle{./IEEEtran}
\bibliography{./IEEEabrv,./ControllableVideoGeneration.bib}

\begin{thebibliography}{10}
\providecommand{\url}[1]{#1}
\csname url@samestyle\endcsname
\providecommand{\newblock}{\relax}
\providecommand{\bibinfo}[2]{#2}
\providecommand{\BIBentrySTDinterwordspacing}{\spaceskip=0pt\relax}
\providecommand{\BIBentryALTinterwordstretchfactor}{4}
\providecommand{\BIBentryALTinterwordspacing}{\spaceskip=\fontdimen2\font plus
\BIBentryALTinterwordstretchfactor\fontdimen3\font minus
  \fontdimen4\font\relax}
\providecommand{\BIBforeignlanguage}[2]{{%
\expandafter\ifx\csname l@#1\endcsname\relax
\typeout{** WARNING: IEEEtran.bst: No hyphenation pattern has been}%
\typeout{** loaded for the language `#1'. Using the pattern for}%
\typeout{** the default language instead.}%
\else
\language=\csname l@#1\endcsname
\fi
#2}}
\providecommand{\BIBdecl}{\relax}
\BIBdecl

\bibitem{takenakaGuidingVideoPrediction2023}
P.~Takenaka, J.~Maucher, and M.~F. Huber, ``Guiding {{Video Prediction}} with
  {{Explicit Procedural Knowledge}},'' in \emph{Proceedings of the
  {{IEEE}}/{{CVF International Conference}} on {{Computer Vision}} ({{ICCV}})
  {{Workshops}}}, Oct. 2023, pp. 1084--1092.

\bibitem{takenakaViProEnablingControlling2024}
------, ``{{ViPro}}: {{Enabling}} and~{{Controlling Video Prediction}}
  for~{{Complex Dynamical Scenarios Using Procedural Knowledge}},'' in
  \emph{Neural-{{Symbolic Learning}} and {{Reasoning}}}, T.~R. Besold,
  A.~{d'Avila Garcez}, E.~{Jimenez-Ruiz}, R.~Confalonieri, P.~Madhyastha, and
  B.~Wagner, Eds.\hskip 1em plus 0.5em minus 0.4em\relax Cham: Springer Nature
  Switzerland, 2024, pp. 62--83.

\bibitem{radfordImprovingLanguageUnderstanding}
A.~Radford, K.~Narasimhan, T.~Salimans, and I.~Sutskever, ``Improving
  {{Language Understanding}} by {{Generative Pre-Training}}.''

\bibitem{marcusRebootingAIBuilding2019}
G.~Marcus and E.~Davis, \emph{Rebooting {{AI}}: {{Building Artificial
  Intelligence We Can Trust}}}.\hskip 1em plus 0.5em minus 0.4em\relax USA:
  Pantheon Books, 2019.

\bibitem{mirzadehGSMSymbolicUnderstandingLimitations2024}
I.~Mirzadeh, K.~Alizadeh, H.~Shahrokhi, O.~Tuzel, S.~Bengio, and M.~Farajtabar,
  ``{{GSM-Symbolic}}: {{Understanding}} the {{Limitations}} of {{Mathematical
  Reasoning}} in {{Large Language Models}},'' Oct. 2024.

\bibitem{duanGTBenchUncoveringStrategic2024}
J.~Duan, R.~Zhang, J.~Diffenderfer, B.~Kailkhura, L.~Sun, E.~{Stengel-Eskin},
  M.~Bansal, T.~Chen, and K.~Xu, ``{{GTBench}}: {{Uncovering}} the {{Strategic
  Reasoning Limitations}} of {{LLMs}} via {{Game-Theoretic Evaluations}},''
  Jun. 2024.

\bibitem{vonruedenInformedMachineLearning2023}
L.~{von Rueden}, S.~Mayer, K.~Beckh, B.~Georgiev, S.~Giesselbach, R.~Heese,
  B.~Kirsch, J.~Pfrommer, A.~Pick, R.~Ramamurthy, M.~Walczak, J.~Garcke,
  C.~Bauckhage, and J.~Schuecker, ``Informed {{Machine Learning}} -- {{A
  Taxonomy}} and {{Survey}} of {{Integrating Prior Knowledge}} into {{Learning
  Systems}},'' \emph{IEEE Transactions on Knowledge and Data Engineering},
  vol.~35, no.~1, pp. 614--633, Jan. 2023.

\bibitem{finnUnsupervisedLearningPhysical2016}
C.~Finn, I.~Goodfellow, and S.~Levine, ``Unsupervised {{Learning}} for
  {{Physical Interaction}} through {{Video Prediction}},'' in \emph{Advances in
  {{Neural Information Processing Systems}}}, vol.~29.\hskip 1em plus 0.5em
  minus 0.4em\relax Curran Associates, Inc., 2016.

\bibitem{fragkiadakiLearningVisualPredictive2016}
K.~Fragkiadaki, P.~Agrawal, S.~Levine, and J.~Malik, ``Learning {{Visual
  Predictive Models}} of {{Physics}} for {{Playing Billiards}},'' Jan. 2016.

\bibitem{fraccaroDisentangledRecognitionNonlinear2017}
M.~Fraccaro, S.~Kamronn, U.~Paquet, and O.~Winther, ``A {{Disentangled
  Recognition}} and {{Nonlinear Dynamics Model}} for {{Unsupervised
  Learning}},'' in \emph{Advances in {{Neural Information Processing
  Systems}}}, vol.~30.\hskip 1em plus 0.5em minus 0.4em\relax Curran
  Associates, Inc., 2017.

\bibitem{jannerReasoningPhysicalInteractions2018}
M.~Janner, S.~Levine, W.~T. Freeman, J.~B. Tenenbaum, C.~Finn, and J.~Wu,
  ``Reasoning {{About Physical Interactions}} with {{Object-Oriented
  Prediction}} and {{Planning}},'' in \emph{International {{Conference}} on
  {{Learning Representations}}}, Sep. 2018.

\bibitem{kandukuriPhysicalRepresentationLearning2022}
R.~K. Kandukuri, J.~Achterhold, M.~Moeller, and J.~Stueckler, ``Physical
  {{Representation Learning}} and {{Parameter Identification}} from {{Video
  Using Differentiable Physics}},'' \emph{International Journal of Computer
  Vision}, vol. 130, no.~1, pp. 3--16, Jan. 2022.

\bibitem{jaquesPhysicsasInverseGraphicsUnsupervisedPhysical2019}
M.~Jaques, M.~Burke, and T.~Hospedales, ``Physics-as-{{Inverse-Graphics}}:
  {{Unsupervised Physical Parameter Estimation}} from {{Video}},'' in
  \emph{International {{Conference}} on {{Learning Representations}}}, Sep.
  2019.

\bibitem{murthyGradSimDifferentiableSimulation2020}
J.~K. Murthy, M.~Macklin, F.~Golemo, V.~Voleti, L.~Petrini, M.~Weiss,
  B.~Considine, J.~{Parent-L{\'e}vesque}, K.~Xie, K.~Erleben, L.~Paull,
  F.~Shkurti, D.~Nowrouzezahrai, and S.~Fidler, ``{{gradSim}}:
  {{Differentiable}} simulation for system identification and visuomotor
  control,'' in \emph{International {{Conference}} on {{Learning
  Representations}}}, Oct. 2020.

\bibitem{liangENVIDRImplicitDifferentiable2023}
R.~Liang, H.~Chen, C.~Li, F.~Chen, S.~Panneer, and N.~Vijaykumar, ``{{ENVIDR}}:
  {{Implicit Differentiable Renderer}} with {{Neural Environment Lighting}},''
  in \emph{Proceedings of the {{IEEE}}/{{CVF International Conference}} on
  {{Computer Vision}}}, 2023, pp. 79--89.

\bibitem{locatelloObjectCentricLearningSlot2020}
F.~Locatello, D.~Weissenborn, T.~Unterthiner, A.~Mahendran, G.~Heigold,
  J.~Uszkoreit, A.~Dosovitskiy, and T.~Kipf, ``Object-{{Centric Learning}} with
  {{Slot Attention}},'' in \emph{Advances in {{Neural Information Processing
  Systems}}}, vol.~33.\hskip 1em plus 0.5em minus 0.4em\relax Curran
  Associates, Inc., 2020, pp. 11\,525--11\,538.

\bibitem{kipfConditionalObjectCentricLearning2022}
T.~Kipf, G.~F. Elsayed, A.~Mahendran, A.~Stone, S.~Sabour, G.~Heigold,
  R.~Jonschkowski, A.~Dosovitskiy, and K.~Greff, ``Conditional {{Object-Centric
  Learning}} from {{Video}},'' in \emph{International {{Conference}} on
  {{Learning Representations}}}, Jan. 2022.

\bibitem{elsayedSAViEndtoEndObjectCentric2022}
G.~F. Elsayed, A.~Mahendran, S.~van Steenkiste, K.~Greff, M.~C. Mozer, and
  T.~Kipf, ``{{SAVi}}++: {{Towards End-to-End Object-Centric Learning}} from
  {{Real-World Videos}},'' in \emph{Advances in {{Neural Information Processing
  Systems}}}, Oct. 2022.

\bibitem{wuSlotFormerUnsupervisedVisual2023}
Z.~Wu, N.~Dvornik, K.~Greff, T.~Kipf, and A.~Garg, ``{{SlotFormer}}:
  {{Unsupervised Visual Dynamics Simulation}} with {{Object-Centric Models}},''
  in \emph{The {{Eleventh International Conference}} on {{Learning
  Representations}}}, Feb. 2023.

\bibitem{zhangUnreasonableEffectivenessDeep2018}
R.~Zhang, P.~Isola, A.~A. Efros, E.~Shechtman, and O.~Wang, ``The
  {{Unreasonable Effectiveness}} of {{Deep Features}} as a {{Perceptual
  Metric}},'' in \emph{2018 {{IEEE}}/{{CVF Conference}} on {{Computer Vision}}
  and {{Pattern Recognition}}}.\hskip 1em plus 0.5em minus 0.4em\relax Salt
  Lake City, UT: IEEE, Jun. 2018, pp. 586--595.

\bibitem{choLearningPhraseRepresentations2014}
K.~Cho, B.~{van Merri{\"e}nboer}, C.~Gulcehre, D.~Bahdanau, F.~Bougares,
  H.~Schwenk, and Y.~Bengio, ``Learning {{Phrase Representations}} using {{RNN
  Encoder}}--{{Decoder}} for {{Statistical Machine Translation}},'' in
  \emph{Proceedings of the 2014 {{Conference}} on {{Empirical Methods}} in
  {{Natural Language Processing}} ({{EMNLP}})}, A.~Moschitti, B.~Pang, and
  W.~Daelemans, Eds.\hskip 1em plus 0.5em minus 0.4em\relax Doha, Qatar:
  Association for Computational Linguistics, Oct. 2014, pp. 1724--1734.

\bibitem{leguenDisentanglingPhysicalDynamics2020}
V.~Le~Guen and N.~Thome, ``Disentangling {{Physical Dynamics From Unknown
  Factors}} for {{Unsupervised Video Prediction}},'' in \emph{2020
  {{IEEE}}/{{CVF Conference}} on {{Computer Vision}} and {{Pattern
  Recognition}} ({{CVPR}})}.\hskip 1em plus 0.5em minus 0.4em\relax Seattle,
  WA, USA: IEEE, Jun. 2020, pp. 11\,471--11\,481.

\bibitem{wangPredRNNRecurrentNeural2023}
Y.~Wang, H.~Wu, J.~Zhang, Z.~Gao, J.~Wang, P.~S. Yu, and M.~Long,
  ``{{PredRNN}}: {{A Recurrent Neural Network}} for {{Spatiotemporal Predictive
  Learning}},'' \emph{IEEE Transactions on Pattern Analysis and Machine
  Intelligence}, vol.~45, no.~2, pp. 2208--2225, Feb. 2023.

\bibitem{donaPDEDrivenSpatiotemporalDisentanglement2021}
J.~Don{\`a}, J.-Y. Franceschi, S.~Lamprier, and P.~Gallinari, ``{{PDE-Driven
  Spatiotemporal Disentanglement}},'' in \emph{International {{Conference}} on
  {{Learning Representations}}}, Jan. 2021.

\bibitem{wuSlotDiffusionObjectCentricGenerative2023}
Z.~Wu, J.~Hu, W.~Lu, I.~Gilitschenski, and A.~Garg, ``{{SlotDiffusion}}:
  {{Object-Centric Generative Modeling}} with {{Diffusion Models}},'' in
  \emph{Thirty-Seventh {{Conference}} on {{Neural Information Processing
  Systems}}}, Nov. 2023.

\end{thebibliography}

\end{document}